\documentclass[sn-mathphys,Numbered]{sn-jnl}
\usepackage{hyperref}
\usepackage{listings}%
\usepackage{graphicx}%
\usepackage{xcolor}%
\usepackage{multirow}%
\usepackage{amsmath,amssymb,amsfonts}%
\usepackage{amsthm}%
\usepackage{mathrsfs}%
\usepackage[title]{appendix}%
\usepackage{textcomp}%
\usepackage{manyfoot}%
\usepackage{booktabs}%
\usepackage{algorithm}%
\usepackage{algorithmicx}%
\usepackage{algpseudocode}%

\usepackage[scaled=0.85]{beramono}
\usepackage[caption=false]{subfig}
  \graphicspath{{./figures/}}
\usepackage{float}

\usepackage{tikz}
\usetikzlibrary{quantikz2}
\usepackage{forest}

\title[article-title]{A quantum k-nearest neighbors implementation using magnitude-aware comparisons}
\author*[1]{\fnm{Christopher} \sur{Rector}}\email{chrisrector14@gmail.com}
\author[1]{\fnm{Jiangwen} \sur{Sun}}\email{jsun@odu.edu}

\affil[1]{\orgdiv{Computer Science}, \orgname{Old Dominion University}, \orgaddress{\street{1 Old Dominion University}, \city{Norfolk}, \postcode{23529}, \state{VA}, \country{United States of America}}}

\begin{document}

\abstract{
    Much attention has recently been devoted to the k-nearest neighbors (KNN) algorithm as a case study for quantum computing algorithms. Several algorithms have devised methods to compute a similarity metric to solve Quantum KNN (QKNN). This paper presents a practical quantum encoding technique that enables Euclidean distance calculation between vectors with different $L2$
    norms by appending a single component to each vector, then uses it in the QKNN problem. It is arguably simpler to understand and implement than comparable methods and uses fewer qubits. Further, an implementation of this algorithm is provided and is used to benchmark it against a recent Euclidean-distance QKNN method on standard labeled datasets. In addition to demonstrating proper formulation, the results show marked improvement over similar implementations during simulation with realistic noise due to the use of fewer qubits and thus more coherent circuits.
}


\maketitle

\section{Introduction}

Machine learning is a popular solution to many otherwise challenging problems spanning various disciplines. A nascent subset of these solutions use quantum computers as an alternate execution medium for well-known machine learning algorithms. Since Shor's solutions \cite{Shor1995PolynomialTimeAF} to the factoring, discrete-log, and period-finding problems, researchers have proposed several algorithms that could achieve quantum supremacy \cite{Preskill2012QuantumCA} over traditional computers. Quantum computing has shown promising improvements to classical algorithms, including proofs of quadratic \cite{Grover1997QuantumMH} and exponential \cite{Shor1995PolynomialTimeAF} computational speedups. This potential for speeding up classical algorithms provides an impetus to re-invent these algorithms, which were designed for classical computers, into implementations that can be executed on quantum computers. The field of machine learning has not been excluded - significant progress has been made in methods ranging from quantum annealing \cite{Venturelli2018ReverseQA} to convolutional neural networks \cite{Hur2021QuantumCN}.

Recent advancements in Quantum Machine Learning (QML) include theoretical or practical implementations of quantum Support Vector Machines (SVMs) \cite{Simon2023OnNQ, Mohanty2023SolvingTV}, k-means \cite{Wu2021QuantumKA, Ohno2022AQA}, and Quantum K-Nearest Neighbors (QKNN) \cite{Wiebe2014QuantumNA}. The K-Nearest Neighbors (KNN) algorithm \cite{fix1985discriminatory}, in which test samples are iteratively assigned labels based on their similarity to pre-labeled training samples, has seen significant efforts toward developing a competitive quantum variant. A detailed review will follow in Section \ref{sec:background}, but some outstanding issues remain to be addressed in the present literature. QKNN implementations use many metrics to evaluate similarity, including cosine similarity \cite{Afham2020QuantumKN, LunaMrquez2025kNNCU, MaldonadoRomo2024QuantumKN}, Mahalanobis distance \cite{Gao2022QuantumKN}, and fidelity \cite{Ma2021QuantumAF}. Euclidean distance has also been used \cite{Yu2020}. Wiebe et al. \cite{Wiebe2014QuantumNA} compute it coherently for normalized vectors, Schuld et al. \cite{Schuld2017DistanceBased} and Berti et al. \cite{Berti2024RoleOE} estimate it with an interference circuit on unit-norm inputs, Lloyd et al. \cite{Lloyd2013QuantumAF} recover it for unnormalized vectors whose norms are supplied separately, and Zardini et al. \cite{Zardini2023AQK} append norm-dependent components that roughly double the dimension of each encoded vector. These approaches either discard the magnitudes of the vectors, need the norms as separate inputs, or substantially enlarge the encoded vectors. This report shares the same professional sentiment as Zardini et al. \cite{Zardini2023AQK} that few published algorithms are practical in the near term. 

This report presents a novel approach to QKNN that addresses these concerns and can be implemented on quantum computers without the use of oracles or other such abstractions. It encodes real-valued vectors with different $L2$ norms by appending a single component to each vector, so that meaningful magnitude comparisons between the original vectors can be retrieved by measuring the quantum state. It improves on previous implementations that use Euclidean distance comparisons by using a much less complicated encoding technique and fewer qubits.

In brief, Section \ref{sec:background} presents relevant background information; Section \ref{sec:algorithm} presents a novel algorithm based on a new quantum encoding technique, along with its basic implementation; Section \ref{sec:experiments} describes the experimental setup and results; Section \ref{sec:conclusions} provides closing remarks and potential future work.

\section{Background} \label{sec:background}

This section describes some relevant background information in quantum computing, current QKNN algorithms, and the problem with using current encoding techniques to compare real-valued vectors with different $L2$ norms in quantum computers.

\subsection{Encoding Methods in Quantum Computing}

Encoding meaningful values into Gate Based Quantum Computers (GBQCs) is still a topic of active research. The following is a review of three prevalent encoding methods. The first, chronologically and in its simplicity, is binary encoding. Familiarity with representing base 2 numbers in classical computers is assumed. Algorithms such as Grover's \cite{Grover1997QuantumMH} use quantum states to represent integers and then apply gates or oracles to solve a problem. Other uses of quantum binary encoding appear in algorithms that implement binary arithmetic on GBQCs, such as this quantum image processing algorithm \cite{Yuan2017QuantumIF}. However, the operations that emerge from these methods use an impractical quantity of qubits for any quantum computing hardware expected in the remote future due to decoherence \cite{decoherence}.

An alternative approach is to encode information in the phase of qubits. This approach enabled Shor to solve his factoring problem \cite{Shor1995PolynomialTimeAF}. Encoding data in the phase of a qubit leverages useful arithmetic properties, such as the addition of exponents. However, phase encoding and inverse quantum Fourier transform subcircuits add more complexity than necessary for computing Euclidean distance, as it only requires addition and multiplication operations and involves no cyclical components.

The approach described in this paper uses amplitude encoding. Given a unit vector $\vec{x} \in \mathbb{R}^N$, amplitude encoding can be described by Equation~\ref{eq:amp}, in which $n = \lceil \log_2(N) \rceil$ and $\vec{x}$ is padded with zeros to length $2^n$.

\begin{equation}
\label{eq:amp}
    U^{(AE)}_x: |0\rangle ^{\otimes n} \rightarrow \sum_{j=0}^{N-1} x_j |j\rangle
\end{equation}

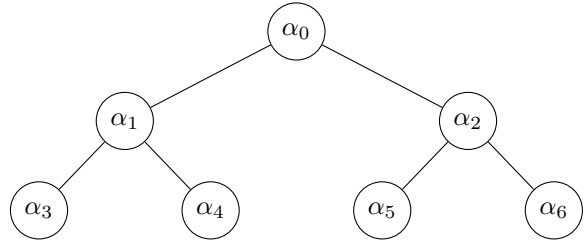
\begin{figure}[H]
    \centering
    {\normalsize
    \begin{forest}
        for tree={s sep=15mm}
        [$\alpha_0$,circle,draw
            [$\alpha_1$,circle,draw
                [$\alpha_3$,circle,draw][$\alpha_4$,circle,draw]
            ]
            [$\alpha_2$,circle,draw
                [$\alpha_5$,circle,draw][$\alpha_6$,circle,draw]
            ]
        ]
    \end{forest}
    }
    \caption{An angle tree to encode a vector into the amplitudes of quantum states}
    \label{fig:angle_tree}
\end{figure}

As described by Araujo et al. \cite{Araujo2020ADA}, this circuit can be implemented with a series of rotation gates around the $y$ axis. Defined as in Nielsen's book \cite{Nielsen2011QuantumCA}, these are known colloquially as $R_y$ gates. The method first generates a tree of angles, as seen in Figure~\ref{fig:angle_tree}.

\begin{figure}[H]
    \centering
    {\tiny
    \begin{quantikz}
    \lstick{\ket{0}} &  &  &  & \gate{R_Y(\alpha_3)} & \gate{R_Y(\alpha_4)} & \gate{R_Y(\alpha_5)} & \gate{R_Y(\alpha_6)} \\
    \lstick{\ket{0}} &  & \gate{R_Y(\alpha_1)} & \gate{R_Y(\alpha_2)} & \octrl{-1} & \ctrl{-1} & \octrl{-1} & \ctrl{-1} \\
    \lstick{\ket{0}} & \gate{R_Y(\alpha_0)} & \octrl{-1} & \ctrl{-1} & \octrl{-1} & \octrl{-1} & \ctrl{-1} & \ctrl{-1}
    \end{quantikz}}
    \caption{A quantum circuit which encodes an 8-element, real-valued vector}
    \label{fig:state_prep}
\end{figure}
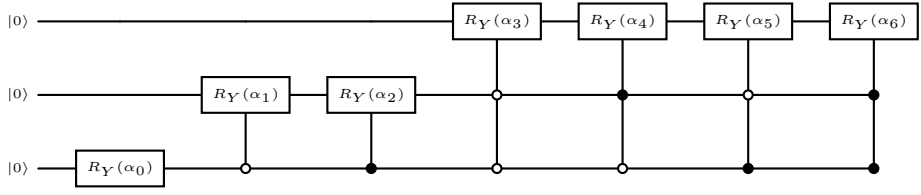

Then, these angles are used as inputs to the controlled-$R_y$ gates, as shown in the circuit in Figure \ref{fig:state_prep}. To encode an $N$ element vector, $O(N)$ classical computations are required to  find the $N - 1$ angles and $N - 1$ quantum gates are required. In reality, gate decomposition will likely be required to convert these abstract gates, to include the multiple control qubits, to those of the universal gate set available on the quantum hardware used to execute the circuit. There are a variety of decomposition methods available, and each has its own strengths and weaknesses. This article \cite{Zindorf2024EfficientIO} provides a more thorough discussion of the subject.

Thus, to encode $p$ vectors in quantum states, this method would require a circuit depth of $O(Np)$ and $\lceil \log_2(p) \rceil + \lceil \log_2(N) \rceil$ qubits. Additionally, Giovannetti et al. \cite{Giovannetti2007QuantumRA} have proposed Quantum Random Access Memory (QRAM) that would be able to accomplish the same in $O(N\log_2(p))$ steps. However, it introduces a linear increase in the number of qubits necessary \cite{Phalak2023QuantumRA} and is still emerging as a practical method of encoding classical data into quantum computers.

\subsection{Quantum k-Nearest Neighbors} \label{sec:qknn_background}

Given a set of labeled features $\mathbf{U} = \{\vec{u}_1, ..., \vec{u}_p\}$ with labels $\vec{y} = \{y_1, ..., y_p\}$ and a test set of features $\mathbf{V} = \{\vec{v}_1, ..., \vec{v}_q\}$, KNN aims to provide labels to each element of $\mathbf{V}$ based on the elements of $\mathbf{U}$. This is done iteratively for each element $\vec{v}_j \in \mathbf{V}$ by finding the subset $\mathbf{U}_k \subset \mathbf{U}$ of $k$ elements described by Definition (\ref{eq:kset}).

\begin{equation}
\label{eq:kset}
    \{\vec{u}_l \quad | \quad d(\vec{v}_j, \vec{u}_l) \le d(\vec{v}_j, \vec{u}_r), \quad \forall \vec{u}_l \in U_k, \quad \forall \vec{u}_r \in U \backslash U_k\}
\end{equation}

$d(\cdot, \cdot)$ is a relevant similarity metric that is discussed in further detail below. The test feature $\vec{v}_j$ adopts the label that appears the most frequently in the set of labels $y_k$ corresponding to the elements of the set $\mathbf{U}_k$.

Many metrics have been proposed for the function $d(\cdot, \cdot)$ in the context of QKNN. One method \cite{Ruan2017QuantumAF} uses a binary adder to find the Hamming distance between input bit strings. This requires an input string register and a difference register of the same size, which quickly becomes inefficient. An image classification QKNN \cite{Zhou2021QuantumKI} was proposed that used the NEQR \cite{Zhang2013NEQRAN} encoding to find the Hamming distance between image representations, then used Grover's algorithm \cite{Grover1997QuantumMH} to select a minimum difference. Another binary-encoded Hamming distance implementation \cite{Li2021QuantumKN} uses an oracle to perform QAEA \cite{Brassard2000QuantumAA} to find the features with minimum distances.

Cosine similarity has been a metric commonly used in KNN. It has taken the form of its square modulus, fidelity, in QKNN. The fidelity $F(\vec{u}, \vec{v})$ is defined as in Equation \ref{eq:fidelity}.

\begin{equation}
\label{eq:fidelity}
    F(\vec{u}, \vec{v}) = |\langle u | v \rangle|^2
\end{equation}

All of these methods \cite{Wiebe2014QuantumNA, Afham2020QuantumKN, LunaMrquez2025kNNCU, Ma2021QuantumAF, Hai2022NewAO, MaldonadoRomo2024QuantumKN, Ronggon2026QKNNNQ} apply some form of amplitude encoding, followed by a controlled swap test. This leaves the squared inner product shown in Equation \ref{eq:fidelity}.

Whether this calculation is performed on a GBQC or classical computer, this measurement is not appropriate for all applications. If the magnitude is pertinent to the similarity of two vectors, then cosine similarity can produce undesirable results. Consider two vectors, $\vec{u}$, $\vec{v} \in \mathbb{R}^n$, such that $\vec{v} = \alpha \vec{u}$ and $\alpha > 0$.

Now consider that all vectors amplitude-encoded into a quantum computer must have $L2$ norms of 1. The normalized versions of $\vec{u}$ and $\vec{v}$ become identical, as shown in Equation \ref{eq:norms}.

\begin{equation}
\label{eq:norms}
    \frac{\vec{u}}{\|\vec{u}\|} = \frac{\vec{v}}{\|\vec{v}\|} = \frac{\alpha \vec{u}}{\alpha \sqrt{\sum_{j=1}^n {u_j^2 }}} = \frac{\vec{u}}{\sqrt{\sum_{j=1}^n {u_j^2 }}} 
\end{equation}

The inner product of these vectors will always be unity, as demonstrated in Equation \ref{eq:alpha_norm}.

\begin{equation}
\label{eq:alpha_norm}
    \langle \frac{\vec{u}}{\|\vec{u}\|} | \frac{\vec{v}}{\| \vec{v} \|} \rangle = \frac{\sum_{j=1}^n {u_j^2 }}{\sum_{j=1}^n {u_j^2 }} = 1
\end{equation}

If $\alpha \approx 0$, then $\vec{v}$ is approximately $\vec{0}$ with some noise. It may not be desirable to consider $\vec{u}$ to be similar to $\vec{v}$ in this case. Additionally, there are some issues with cosine similarity and normalized Euclidean distance results converging when comparing samples in high-dimensional feature spaces \cite{Tessari2024SurpassingCS}.

Another approach \cite{Gao2022QuantumKN} encodes a covariance matrix into a gate operator using a Hamiltonian to calculate the Mahalanobis distance between features. It then uses QAEA to find the closest features. Mahalanobis distance can be reduced to Euclidean distance using the identity matrix as the covariance matrix. Zardini et al. \cite{Zardini2023AQK} propose a metric based on Euclidean distance using two new encoding methods which they call extension and translation. Their approach is similar to the one presented in this report: rather than normalizing input vectors to an $L2$ norm of $1$, they rescale each feature to a fixed range and append norm-dependent components so that every encoded vector has unit norm. However, it uses a complicated conditional encoding scheme in which a vector with $N$ features becomes $2N + 3$ or $2N + 4$ components.

\subsection{Comparing vectors with different $L2$ norms}

A goal of a few of the algorithms discussed in Section \ref{sec:qknn_background} is to compute the squared Euclidean distance between two vectors $\vec{u}, \vec{v} \in \mathbb{R}^n$, as defined in Equation \ref{eq:euclidean}.

\begin{equation}
\label{eq:euclidean}
    d_E^2(\vec{u}, \vec{v}) = \|\vec{u} - \vec{v}\|^2
\end{equation}

Some of these approaches are able to find a value proportional to it when constrained such that $\|\vec{u}\| = \|\vec{v}\| = 1$. However, not all preprocessing techniques commonly used in machine learning accomplish this. Min-max scaling, as defined in Equation \ref{eq:minmax}, and z-score scaling, as defined in Equation \ref{eq:zscore}, do not achieve this constraint. $\mu$ is the sample mean and $\sigma$ is the standard deviation. 

\begin{equation}
\label{eq:minmax}
    x' = \frac{x - x_{min}}{x_{max} - x_{min}}
\end{equation}

\begin{equation}
\label{eq:zscore}
    z = \frac{x - \mu}{\sigma}
\end{equation}

Some preprocessing techniques do scale feature vectors by setting their $L2$ norms equal to 1. However, this leaves the resulting vectors in a similar situation to the cosine similarity discussed in Section \ref{sec:qknn_background}. Only the feature vector's orientation is considered, but not its magnitude. In fairness, some preprocessing standards exist that first find the min-max or z-score representation and then normalize the vectors to have $L2$ norms of 1, but these are not used universally as of the time of this writing.

All input vectors must have an $L2$ norm equal to $1$ to encode them into a GBQC. Define $\vec{x}_{norm}$ as the unit-normalized version of vector $\vec{x}$ such that $\|\vec{x}_{norm}\|_2 = 1$. A quantum circuit can be created such that the probabilities of measuring certain states are proportional to Euclidean distance between these normalized vectors, as defined in Equation \ref{eq:normed_euclidean}. The specifics of the circuit shall be discussed in Section \ref{sec:quclidean}.

\begin{equation}
\label{eq:normed_euclidean}
    d_E^2(\vec{u}_{norm}, \vec{v}_{norm}) = \sum_{j=1}^n{(\frac{u_j}{||\vec{u}||} - \frac{v_j}{||\vec{v}||})^2}
\end{equation}

\begin{figure}[htb]
    \centering
    \includegraphics[width=0.8\textwidth]{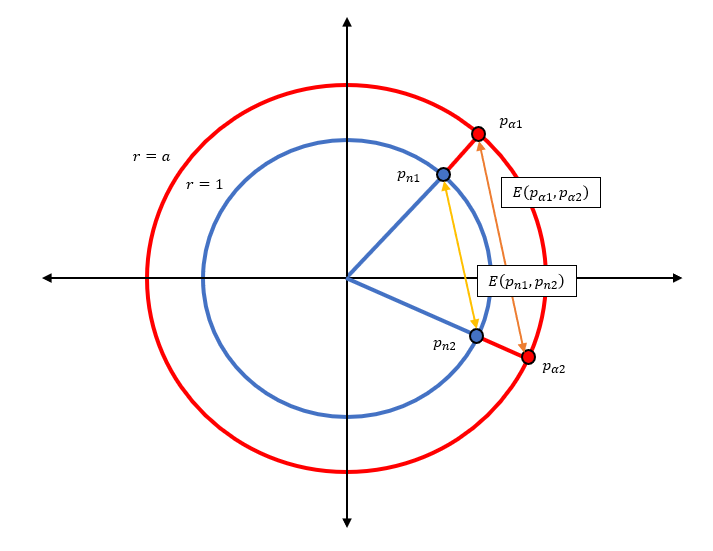}
    \caption{Points $p_{\alpha1}$ and $p_{\alpha 2}$ normalized as $p_{n1}$ and $p_{n2}$ to fit the encoding constraints of GBQCs.}
    \label{fig:same_norm_circle}
\end{figure}

It may appear that this method has successfully found the original Euclidean distance. However, there is a severe limitation to this method. It requires $||\vec{u}|| = ||\vec{v}||$, $\forall u \in U, \forall v \in V$, as shown graphically in Figure \ref{fig:same_norm_circle}. 

If $||\vec{v}|| = ||\vec{u}||$, then Equation \ref{eq:same_norm_factored} can be obtained by factoring out the common denominator. Thus, the original Euclidean distance is proportional to the normalized Euclidean distance via the relationship shown in Equation \ref{eq:same_norm_relation}.

\begin{equation}
\label{eq:same_norm_factored}
    d_E^2(\vec{u}_{norm}, \vec{v}_{norm}) = \frac{1}{||\vec{u}||^2}\sum_{j=1}^n{(u_j - v_j)^2}
\end{equation}

\begin{equation}
\label{eq:same_norm_relation}
    d_E^2(\vec{u}, \vec{v}) = ||\vec{u}||^2 d_E^2(\vec{u}_{norm}, \vec{v}_{norm})
\end{equation}

\begin{figure}[htb]
    \centering
    \includegraphics[width=0.8\textwidth]{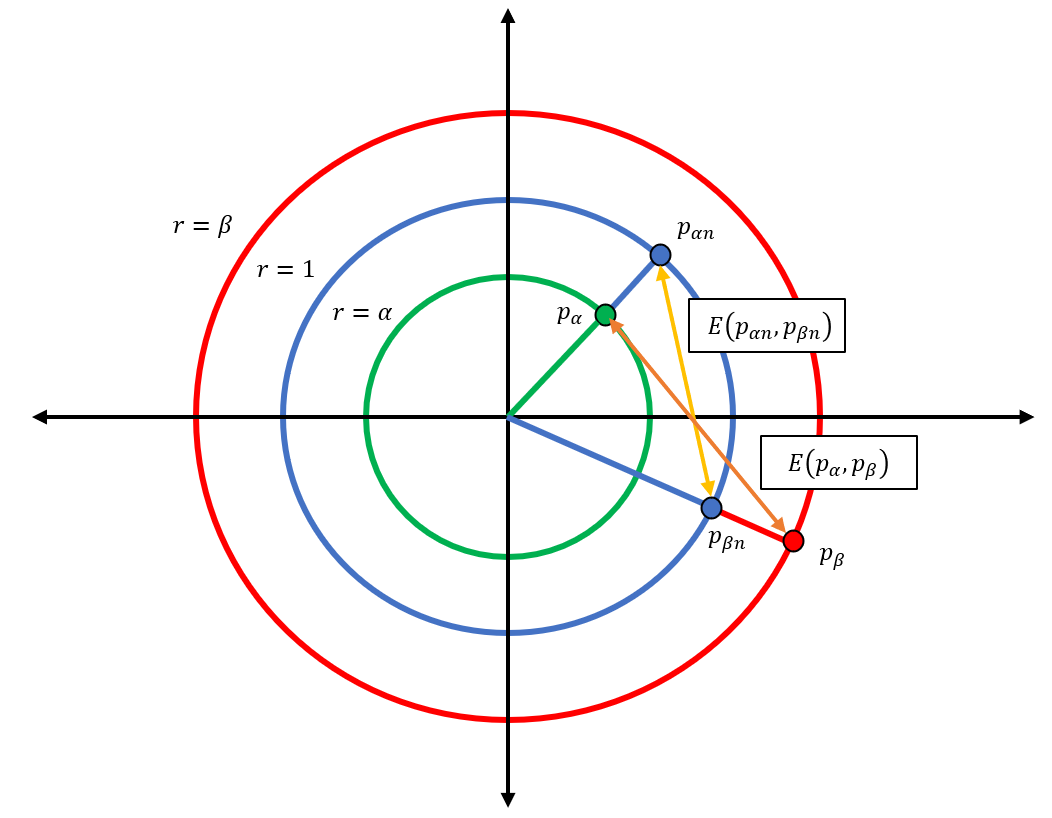}
    \caption{To be encoded into a GBQC, the points $p_\alpha$ and $p_\beta$ must be normalized to $p_{\alpha n}$ and $p_{\beta n}$, respectively.}
    \label{fig:norm_space_circles}
\end{figure}

Now consider the case in which $||\vec{u}|| \ne ||\vec{v}||$. This is shown graphically in Figure \ref{fig:norm_space_circles}. Once normalized to meet the constraints of GBQCs, the result still follows from Equation \ref{eq:normed_euclidean}. However, the simplification used to obtain Equation \ref{eq:same_norm_factored} cannot be used here. Further, Equation \ref{eq:diff_norms} shows that the result cannot be linearly related to the Euclidean distance between the original vectors.

\begin{equation}
\label{eq:diff_norms}
    d_E^2(\vec{u}_{norm}, \vec{v}_{norm}) = \frac{1}{{||\vec{u}||^2 ||\vec{v}||^2}}\sum_{j=1}^n{(u_j ||\vec{v}|| - v_j ||\vec{u}||})^2 \ne \alpha d_E^2(\vec{u}, \vec{v})
\end{equation}

Thus, vectors with different $L2$ norms cannot be compared directly using any similar technique.

\section{Algorithm} \label{sec:algorithm}

The preceding approach is not practical unless all vectors $\vec{u} \in \boldsymbol{U}, \vec{v} \in \boldsymbol{V}$ have the same $L2$ norm - that is, $||\vec{u}_j||_2 = ||\vec{v}_l||_2, \forall j, l$. This is often not the case for applications in science and engineering. Therefore, the vectors under consideration must be modified to exist in the same norm space. Section \ref{sec:map} describes a simple, yet novel, encoding method that modifies all input vectors to have the same $L2$ norm. This allows the use of Equation~\ref{eq:same_norm_relation} to recover exact Euclidean distances between the original vectors upon measuring the circuit in Section \ref{sec:quclidean}. Section \ref{sec:qknn} shows how this method can be used to implement the well-known KNN algorithm.

\subsection{Vector Mapping} \label{sec:map}

This section shows that artificially increasing the dimension of the input vectors can allow retrieval of the Euclidean distances between the original vectors after encoding them into a GBQC. Given a vector $\vec{u} \in \mathbb{R}^n$, select a vector space of size $m$ such that $m = 2^s$, $s \in \mathbb{N}$, and $m > n$. Let a new vector $\vec{w} \in \mathbb{R}^m$ be defined as in Equation \ref{eq:w}.

\begin{equation}
\label{eq:w}
    \vec{w} = \begin{bmatrix}
        u_1 \\
        \vdots \\
        u_n \\
        a_1 \\
        \vdots \\
        a_{m-n}
    \end{bmatrix}
\end{equation}

Let the sum of the squares of the artificial elements $c^2$ be defined as in Equation \ref{eq:d}. The sole purpose of $c$ is to manipulate the value $\|\Vec{w}\|_2$.

\begin{equation}
\label{eq:d}
    c^2 = \sum_{j=1}^{m - n} a_j^2
\end{equation}

Choose the maximum $L2$ norm of all the vectors in sets $\mathbf{U}$ and $\mathbf{V}$, as in Equation~\ref{eq:beta_norm}. The maximum constraint ensures that it is possible to map every $\vec{u}_i, \vec{v}_j$ to a vector $\vec{w}_k$ within the same norm space.

\begin{equation}
\label{eq:beta_norm}
    \beta = \underset{j, l}{\max}\{||\vec{u_j}||, ||\vec{v_l}||\}
\end{equation}

To find $c$, subtract the sum of the squared elements of $\vec{u}$ from the square of the norm $\beta$ as in Equation \ref{eq:d_solve}.

\begin{equation}
\label{eq:d_solve}
    c = \sqrt{\beta^2 - ||\vec{u}||^2}
\end{equation}

\begin{figure}[htb]
    \centering
    \includegraphics[width=0.8\textwidth]{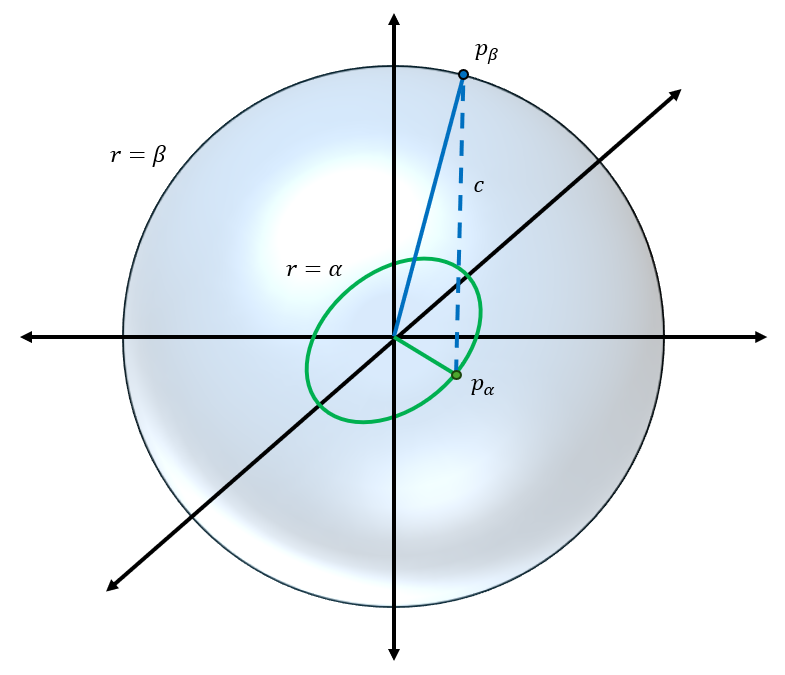}
    \caption{Artificially modifying a point $p_\alpha$ to map it to the surface of a higher dimensional sphere of radius $\beta$.}
    \label{fig:norm_sphere}
\end{figure}

A practical choice for the augmented vector $\vec{w}$ would then be the value $c$ inserted into the $n + 1$ element and the remaining elements set to $0$, as shown in Equation \ref{eq:w_prac}.

\begin{equation}
\label{eq:w_prac}
    \vec{w} = \begin{bmatrix}
        u_1 \\
        \vdots \\
        u_n \\
        c \\
        \vdots \\
        0
    \end{bmatrix}
\end{equation}

These augmented vectors $\vec{w}_i$ can then be substituted for the vectors $\vec{u}_j$ and $\vec{v}_l$ in the conditional amplitude encoding step described later in Equation \ref{eq:encode}. Then, all vectors in both sets will have been mapped to the same norm $\beta$, as depicted graphically by the spaces $S_\alpha$ and $S_\beta$ in the $\mathbb{R}^2 \rightarrow \mathbb{R}^3$ transformation example in Figure \ref{fig:norm_sphere}.

$$ S_\alpha = \{\vec{u} = a \hat{x} + b \hat{y} \quad | \quad a^2 + b^2 \le \beta^2\} $$

$$ S_\beta = \{\vec{u} = a \hat{x} + b \hat{y} + c \hat{z} \quad | \quad a^2 + b^2 + c^2 = \beta^2\} $$

This method of encoding shall be referred to as \textit{map encoding} for the remainder of this report. A similar augmentation, which appends $\sqrt{\beta^2 -||u||_2^2}$ to give a set of vectors a common norm, is used classically to reduce maximum inner product search to nearest-neighbor search \cite{Bachrach2014SpeedingUT}. Let the map encoding procedure be denoted as the transform $T : \vec{x} \rightarrow \vec{y}$ with $\vec{x} \in \mathbb{R}^n$ and $\vec{y} \in \mathbb{R}^m$. Then, let $\vec{w} = T(\vec{u})$ and $\vec{z} = T(\vec{v})$. With $\| \vec{w} \| = \| \vec{z} \| = \beta$, these vectors must be unit-normalized into $\vec{w}_{norm}$ and $\vec{z}_{norm}$, such that $\|\vec{w}_{norm}\|_2=\|\vec{z}_{norm}\|_2 = 1$, so that they are eligible to be encoded into a quantum state.

The next section presents a circuit that can find the sum of squared differences presented in Equation \ref{eq:wz_distance}

\begin{equation}
\label{eq:wz_distance}
    d_E^2(\vec{w}_{norm}, \vec{z}_{norm}) = \sum_{j=1}^m(\frac{w_j}{\beta} - \frac{z_j}{\beta})^2
\end{equation}

 Equation \ref{eq:squared_euclidean} is obtained using the same factoring performed in Equation \ref{eq:same_norm_factored}, 

\begin{equation}
\label{eq:squared_euclidean}
    d_E^2(\vec{w}_{norm}, \vec{z}_{norm}) = \frac{1}{\beta^2}\sum_{j=1}^m(w_j - z_j)^2
\end{equation}

Equation \ref{eq:squared_euclidean} is further disassembled using the definition in Equation \ref{eq:w_prac} to obtain the squared differences of the elements of $\vec{u}$ and $\vec{v}$ in Equation \ref{eq:map_euclidean}.

\begin{equation}
\label{eq:map_euclidean}
    d_E^2(\vec{w}_{norm}, \vec{z}_{norm}) = \frac{1}{\beta^2}\sum_{j=1}^n(u_j - v_j)^2 + \frac{1}{\beta^2}\sum_{l=n + 1}^m (w_l - z_l)^2
\end{equation}

If the values of the second term in Equation \ref{eq:map_euclidean} can be discarded, only the value $d_E^2(\vec{u}_{norm}, \vec{v}_{norm})$ remains in Equation \ref{eq:prop_euclidean}. The next section shows how to obtain a measurement of this value via a GBQC.

\begin{equation}
\label{eq:prop_euclidean}
    d_E^2(\vec{u}_{norm}, \vec{v}_{norm}) = \frac{1}{\beta^2} d_E^2(\vec{u}, \vec{v})
\end{equation}

\subsection{Quantum Euclidean Distance} \label{sec:quclidean}

What follows is a description of a quantum circuit that calculates the Euclidean distance between a test point and all sample points. It builds on the interference circuit of Schuld et al. \cite{Schuld2017DistanceBased}, applied here to map-encoded vectors. First, modify all training set vectors $\vec{u} \in \mathbf{U}$ and test set vectors $\vec{v} \in \mathbf{V}$ following the procedure described in Section \ref{sec:map}, creating the augmented training set vectors $\vec{w} \in \mathbf{W}$ and test set vectors $\vec{z} \in \mathbf{Z}$. For clarity, $\vec{u}, \vec{v} \in \mathbb{R}^N$ and $\vec{w}, \vec{z} \in \mathbb{R}^M$ and $|\mathbf{U}| = |\mathbf{W}| = Q$ and $|\mathbf{V}|$ = $|\mathbf{Z}| = R$. For convenience, $m = \lceil \log_2(M) \rceil$, $q = \lceil \log_2(Q) \rceil$, and $r = \lceil \log_2(R) \rceil$. All augmented vectors have an $L2$ norm of $\beta$. Initialize a circuit with three registers with state $|\phi_0\rangle$ in Equation \ref{eq:init0}. The notation $|\phi\rangle = |q_0\rangle_0 |q_1 \rangle_1 |q_2 \rangle_2$ will be used to track the state of each quantum register throughout the circuit.

\begin{equation}
\label{eq:init0}
    | \phi_0 \rangle = | 0 \rangle_0 | 0 \rangle_1 | 0 \rangle_2
\end{equation}

Apply a Hadamard gate to all registers except the vector data register $|q_2\rangle$ to produce the state $|\phi_1\rangle$ in Equation \ref{eq:H}, which puts the remaining registers into an equal superposition and serves as an address register.

\begin{equation}
\label{eq:H}
    | \phi_1 \rangle = (H \otimes H^{\otimes q} \otimes I^{\otimes m}) |\phi_0\rangle
\end{equation}

$$ | \phi_1 \rangle = \frac{1}{\sqrt{2Q}} \sum_{j=0}^{Q-1}
    (| 0 \rangle_0 + | 1 \rangle_0) \otimes 
    | j \rangle_1 | 0 \rangle_2 $$

With an amplitude encoding unitary $U^{(AE)}_x$ as defined in Equation \ref{eq:amp}, a controlled variant of this unitary is additionally defined as $C_{jl}(U_x^{(AE)})$ and an anti-controlled variant $\tilde{C}_{jl}(U_x^{(AE)})$. When qubit register $|q_0\rangle$ is in the state $|1\rangle$, the controlled gate amplitude encodes $\vec{x}$ into qubit register $l$. The anti-controlled gate amplitude encodes $\vec{x}$ into qubit register $l$ when qubit register $j$ is in the state $|0\rangle$.

These gates can encode the sets of vectors into the vector data register $|q_2\rangle$. The ancillary qubit $|q_0\rangle$ conditionally encodes the vectors of set $\mathbf{W}$ when $|q_0\rangle = |0\rangle$ and the test sample $\vec{z_l}$ when $|q_0\rangle = |1\rangle$. The address register $|q_1\rangle$ is used to encode the data in the appropriate states. This results in state $|\phi_2\rangle$ in Equation \ref{eq:encode}.

\begin{equation}
\label{eq:encode}
    | \phi_2 \rangle = 
    \tilde{C}_{02}(U_{W}^{(AE)}) C_{02}(U_{Z}^{(AE)}) | \phi_1 \rangle
\end{equation}

$$ | \phi_2 \rangle = \frac{1}{\sqrt{2Q}} \sum_{j=0}^{Q-1} \sum_{\alpha = 0}^{M-1}
    (w_{j\alpha} | 0 \rangle_0 + z_{l\alpha} | 1 \rangle_0) \otimes | j \rangle_1 | \alpha \rangle_2 $$

\begin{figure}[H]
\centering
\begin{quantikz}[wire types={q,b,b}]
\lstick{$\ket{0}_0$} & \gate{H} & \octrl{1} & \ctrl{2} & \gate{H} & \meter{} \\
\lstick{$\ket{0}_1$} & \gate{H} & \gate[2]{Encode} &  &  & \meter{} \\
\lstick{$\ket{0}_2$} &  &  & \gate[1]{Encode} &  & \meter{} \\
\end{quantikz}
\caption{The quantum circuit to compute Euclidean distance, with the ancillary qubit $|a\rangle$ in the zeroth quantum register, the address register $|j\rangle$ in the first, and the data $|\alpha\rangle$ in the second.}
\label{fig:encode}
\end{figure}
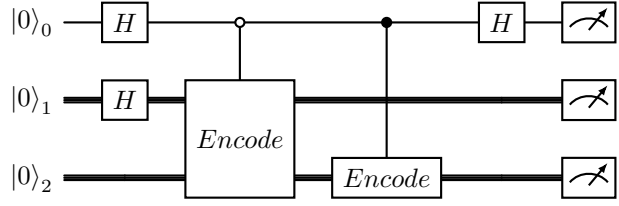

Finally, apply a Hadamard gate to the ancillary qubit $|q_0\rangle$, resulting in state $|\phi_3\rangle$ shown in Equation \ref{eq:decode}. Figure \ref{fig:encode} shows the circuit diagram for this entire procedure.

\begin{equation}
\label{eq:decode}
    | \phi_3 \rangle = 
    (H \otimes I^{\otimes q} \otimes I^{\otimes m}) |\phi_2\rangle 
\end{equation}

$$ |\phi_3 \rangle = \frac{1}{\sqrt{4Q}} \sum_{j=0}^{Q-1} \sum_{\alpha = 0}^{M-1} 
    ((w_{j\alpha} + z_{l\alpha}) | 0 \rangle_0 + 
    (w_{j\alpha} - z_{l\alpha}) | 1 \rangle_0 ) \otimes |j \rangle_1 | \alpha \rangle_2 $$

Define the projection $P_{j \alpha} = |1 \rangle_0 |j \rangle_1 |\alpha \rangle_2 \langle 1|_0 \langle j|_1 \langle \alpha|_2$. Then, the probabilities of measuring this observable are shown in Equation \ref{eq:final_measurement}.

\begin{equation}
\label{eq:final_measurement}
    \langle \phi_3| P_{j\alpha} |\phi_3 \rangle = p(1, j, \alpha) = \frac{(w_{j \alpha} - z_{l \alpha})^2}{4Q}
\end{equation}

To retrieve the Euclidean distances, add the probabilities over the range $\alpha = \{0, ..., N-1\}$ while discarding the probabilities in range $\alpha = \{N, ..., M - 1\}$. Then, multiply by $4Q \beta^2$ to obtain the squared Euclidean distances in Equation \ref{eq:final_euclidean} to match the relationship described in Equation \ref{eq:prop_euclidean}.

\begin{equation}
\label{eq:final_euclidean}
    d_E^2(\vec{u}_j, \vec{v}_l) = 4Q \beta^2 \sum_{\alpha=0}^{N - 1} p(1, j, \alpha) = \beta^2 \sum_{\alpha=0}^{N - 1} (w_{j\alpha} - z_{l\alpha})^2
\end{equation}

\subsection{QKNN} \label{sec:qknn}

This section describes the QKNN algorithm in its entirety. Given the training and test sets $\mathbf{U}$ and $\mathbf{V}$ with $\vec{u}_j, \vec{v}_l \in \mathbb{R}^N$ and label set $\vec{y}$ as described in Section \ref{sec:qknn_background}, perform any desired preprocessing. Then, follow the procedure in Listing \ref{lst:qnn} to assign labels to the test samples in set $\mathbf{V}$. \lstinline{norm_scale} is the procedure described in Section \ref{sec:map}, $H$ is the Hadamard gate transform, \lstinline{encode} is the gate transform described in Equation \ref{eq:encode}, \lstinline{execute_circuit} is the execution and measurement of the quantum circuit, and \lstinline{assign_label} finds the $k$ training features with the lowest summed probabilities $\sum_{\alpha = 0}^{N-1} p(1, j, \alpha)$ as described in Equation \ref{eq:kset}, and returns the associated label that occurs most frequently. The notation $|\gamma\rangle_{jl \alpha}$ as a parameter of a function implies that the function acts on the quantum registers $j$, $l$, and $\alpha$ of the quantum state $|\gamma \rangle$. $|\cdot|$ is the cardinality of a set.

\begin{lstlisting}[escapeinside={(*}{*)}, numbers=left, caption=QKNN algorithm for a GBQC, label=lst:qnn]
    U', V' = norm_scale(U, V)
    (*$m = \lceil \log_2(N + 1) \rceil$*)
    (*$q = \lceil \log_2(|U|) \rceil$*)
    (*$|\psi\rangle = |0\rangle_0 |0\rangle_1^{\otimes q} |0\rangle_2^{\otimes m} $*) 
    (*$|\psi\rangle = H(|\psi\rangle_{01})$*)
    for (*$\vec{u}_j$*) in U':
        (*$|\psi\rangle$*) = encode((*$|\psi\rangle_{012}, \text{ } \vec{u}_j$*))
    for (*$\vec{v}_l$*) in V':
        (*$|\phi\rangle$*) = encode((*$|\psi\rangle_{02}, \text{ } \vec{v}_l$*))
        (*$|\phi\rangle = H(|\phi\rangle_0)$*)
        measurements = execute_circuit((*$|\phi\rangle$*))
        (*$y_l$*) = assign_label((*$\vec{y}$*), measurements)
\end{lstlisting}

The implementation, provided \href{https://github.com/crect500/squib}{here}, uses Qiskit \cite{Qiskit} to construct and simulate the quantum circuits. The \lstinline{assign_label} function does not need to convert the measurements back into distances or unit vectors to determine which features are most similar. This is because these conversions would consist of multiplying all measurements by the same positive, real number.

\subsection{Complexity} \label{sec:complexity}

Let $|\mathbf{U}| = Q$, $|\mathbf{V}| = R$, and $s = Q + R$. We can examine each step of the algorithm to determine the classical computational complexity and the depth and width of the quantum circuit needed to implement the algorithm.

The first step is to determine each element $c$ that maps each vector to the same higher-dimensional surface as described in Section \ref{sec:map}. This requires $O(sN)$  computations.

Currently, amplitude encoding one vector with $N$ elements takes $O(N)$ gates. With $Q$ vectors to encode, this would take $O(QN)$ gates \cite{Araujo2020ADA}. This process would theoretically take $O(\log_2(Q)\log_2(N))$ gates with QRAM. Hadamard gates can be applied in constant depth, $O(1)$.

Once the circuit is executed and the state counts are measured, applications may measure one quantum state or all quantum states. If all states are desired, this requires measuring all $Q$ differences. However, if the results are indexed as needed, each lookup can be performed in $O(1)$ steps.

The entire circuit then has a depth of $O(QN)$, or $O(\log_2(Q)\log_2(N))$ if QRAM is available, with a classical computational complexity of $O(sN)$ for the setup. Each circuit requires $1$ ancillary qubit, $\lceil \log_2(Q) \rceil $ qubits to create address registers for the vectors in $U$, and $\lceil \log_2(N + 1)\rceil$ qubits to encode each vector in $\mathbb{R}^N$. This results in a circuit of width $\lceil \log_2(Q) \rceil + \lceil \log_2(N + 1) \rceil + 1$. $R$ circuits must be executed to run the QKNN algorithm.

Table \ref{tab:complexity} shows a comparison of circuit width, circuit depth, and the number of circuits required to complete the algorithm between a number of QKNN implementations. We use the QRAM-encoded complexities to be consistent with those reported by other papers. $F$ is either $2N + 3$ or $2N + 4$ as defined by Zardini et al. \cite{Zardini2023AQK}, depending on the encoding method.

\begin{table}[h]
    \centering
    \begin{tabular}{|c|c|c|c|c|}
        \hline
        \textbf{Author} & \textbf{Metric} & \textbf{Circuit Depth} & \textbf{Circuit Width} & \textbf{Circuits} \\
        \hline
        Fastovets \cite{Fastovets2019MachineLM} & Fidelity & $O(\log_2(N))$ & $ \lceil \log_2(N) \rceil $ & $QR$ \\
        \hline
        Hai \cite{Hai2022NewAO} & Fidelity & $O(\log^2_2(N))$ & $ O(N) $ & $QR$ \\
        \hline
        Afham \cite{Afham2020QuantumKN} & Fidelity & $O(\log_2(N))$ & $ O(\log_2(QN)) $ & $R$ \\
        \hline
        Ruan \cite{Ruan2017QuantumAF} & Hamming & $O(N^3)$ & $ N(Q + 1) + 1 $ & $R$ \\
        \hline
        Zardini \cite{Zardini2023AQK} & Euclidean & $ O(\log_2(QF) + N)$ & $\lceil \log_2(Q) \rceil + \lceil \log_2(F) \rceil + 2$ & $R$ \\
        \hline
        This paper & Euclidean & $O(\log_2(Q)\log_2(N))$ & $\lceil \log_2(Q) \rceil + \lceil \log_2(N + 1) \rceil + 1$ & $R$ \\
        \hline
    \end{tabular}
    \caption{A comparison of circuit depth and width between quantum KNN algorithms}
    \label{tab:complexity}
\end{table}

\section{Experiments} \label{sec:experiments}

\subsection{Hardware}

All experiments were executed on Intel\textregistered \text{ }Xeon\textregistered \text{ }Gold 6148 CPUs at 2.4GHz with 384GB of RAM, provided by Old Dominion University's Wahab high-performance computing cluster. Quantum computers were not used because of their current limited capabilities and availability to the public.

\subsection{Datasets}

In an effort to provide an accurate comparison with relevant methods, we used the same datasets from the \href{https://archive.ics.uci.edu/}{UCI Machine Learning Repository} that Zardini et al. \cite{Zardini2023AQK} used to evaluate the extension and translation methods. The properties of each dataset are shown in Table \ref{tab:datasets} for quick reference. Although our code is capable of running QKNN on datasets with any number of labels, the datasets used only had binary labels because the previous method's \cite{Zardini2023AQK} code is not set up for multi-class prediction at this time and direct comparison with their results was desired.

\begin{table}[htb]
    \centering
    \begin{tabular}{|c|c|c|c|}
        \hline
        Dataset & Classes & Size & Features \\
        \hline
        \href{https://archive.ics.uci.edu/dataset/53/iris}{Iris - setosa \& versicolor} & 2 & 100 & 4 \\
        \href{https://archive.ics.uci.edu/dataset/53/iris}{Iris - setosa \& virginica} & 2 & 100 & 4 \\
        \href{https://archive.ics.uci.edu/dataset/53/iris}{Iris - versicolor \& virginica} & 2 & 100 & 4 \\
        \href{https://archive.ics.uci.edu/dataset/176/blood+transfusion+service+center}{Transfusion} & 2 & 748 & 4 \\
        \href{https://archive.ics.uci.edu/dataset/212/vertebral+column}{Vertebral column} & 2 & 310 & 6 \\
        \href{https://archive.ics.uci.edu/dataset/39/ecoli}{Ecoli - cp \& im} & 2 & 220 & 7 \\
        \href{https://archive.ics.uci.edu/dataset/42/glass+identification}{Glass - 1 \& 2} & 2 & 146 & 9 \\
        \href{https://archive.ics.uci.edu/dataset/451/breast+cancer+coimbrar}{Breast cancer} & 2 & 116 & 9 \\
        \hline
    \end{tabular}
    \caption{Datasets provided by UCI}
    \label{tab:datasets}
\end{table}

To provide the most accurate comparison, this study uses as similar parameters and metrics as possible. A \textit{k}-fold cross-validation setup was used to split the datasets into test and training sets before running the algorithm. The number of folds chosen was 5 - a default parameter in most machine learning training pipelines. The \textit{k} values of the KNN algorithm for this experiment were set to \textit{3}, \textit{5}, \textit{7}, and \textit{9}. This \textit{k} value is not to be confused with the \textit{k} in \textit{k}-fold cross-validation, as the number of folds is \textit{5} for the entirety of this report. The same k-fold seed was used for all experiments, ensuring that all implementations were provided with the same sets of training and test features. Each data point in the results provided in the following sections represents the average of the results over all folds.

\subsection{Execution}

Two experiments were performed to assess the performance of the QKNN algorithm. Both experiments used a statevector simulator to evolve the quantum state to the desired Euclidean distance used in the KNN algorithm. The first experiment calculated the Euclidean distance using Qiskit's \href{https://qiskit.github.io/qiskit-aer/stubs/qiskit_aer.StatevectorSimulator.html}{StatevectorSimulator} to perform an ideal statevector simulation without noise. The second experiment used a realistic noise model based on the historical performance of IBM's Guadalupe Falcon r4P quantum processor via the \href{https://docs.quantum.ibm.com/api/qiskit-ibm-runtime/fake-provider-fake-guadalupe-v2}{FakeGuadalupeV2} and \href{https://qiskit.github.io/qiskit-aer/tutorials/1_aersimulator.html}{AerSimulator} APIs provided by Qiskit. This mode introduces a restriction of the resolution of the measurements by running the circuit only a given number of times - a quantity commonly referred to as \textit{shots}. Both methods were compared against the KNN algorithm using traditional binary arithmetic to calculate the Euclidean distances, which we hereafter refer to as classical KNN.

\subsection{Metrics}

The accuracy, Jaccard Index (JI), and Average JI (AJI) were recorded for every experiment. If $A$ is the set of actual nearest neighbor features found using classical computation and $P$ is the set of predicted nearest neighbor features, then the $JI$ is defined as $\gamma$ in Equation \ref{eq:ji}.

\begin{equation}
\label{eq:ji}
    JI = \gamma(A, P) = \frac{|A \cap P|}{|A \cup P|}    
\end{equation}

The $AJI$ is computed by iteratively computing the $JI$ for the set of $m$ most similar features, up through the $k$ value, as shown in Equation \ref{eq:aj}. $A_m$ and $P_m$ are the subsets of $A$ and $P$, respectively, containing the $m$ most similar features to the test feature.

\begin{equation}
\label{eq:aj}
    AJI = \frac{1}{k} \sum_{m=1}^k \gamma(A_m, P_m)    
\end{equation}

\subsection{Results} \label{sec:results}

\subsubsection{Algorithm Performance} \label{sec:performance}

The results of the experiments are publicly available \href{https://github.com/crect500/qknn_results}{here}.

\begin{table}[htb]
    \centering
    \resizebox{\textwidth}{!}{%
    \begin{tabular}{|c|ccc|ccc|ccc|ccc|}
        \hline
        \rule{0pt}{4mm} \multirow{2}{19mm}{\textbf{Experiment}} & \multicolumn{3}{c|}{\textbf{k=3}} & \multicolumn{3}{c|}{\textbf{k=5}} & \multicolumn{3}{c|}{\textbf{k=7}} & \multicolumn{3}{c|}{\textbf{k=9}} \\ [1mm]
         & \textbf{$\Delta$ Acc} & \textbf{JI} & \textbf{AJI} & \textbf{$\Delta$ Acc} & \textbf{JI} & \textbf{AJI} & \textbf{$\Delta$ Acc} & \textbf{JI} & \textbf{AJI} & \textbf{$\Delta$ Acc} & \textbf{JI} & \textbf{AJI} \\
        \hline
        \rule{0pt}{4mm} \multirow{2}{19mm}{Setosa-Versicolor} & 0.000 & 0.993 & 0.993 & 0.000 & 0.988 & 0.989  & 0.000 & 0.987  & 0.990 & 0.000 & 0.986 & 0.989  \\ [1mm]
         & $\pm$ 0.000 & $\pm$ 0.013 & $\pm$ 0.014 & $\pm$ 0.000 & $\pm$ 0.010 & $\pm$ 0.012 & $\pm$ 0.000 & $\pm$ 0.014 & $\pm$ 0.013 & $\pm$ 0.000 & $\pm$ 0.016 & $\pm$ 0.013 \\ [1mm]
        \hline
        \rule{0pt}{4mm} \multirow{2}{19mm}{Setosa-Virginica} & 0.000 & 0.997 & 0.994 & 0.001 & 0.988 & 0.992  & 0.000 & 0.989  & 0.992 & 0.001 & 0.984 & 0.990  \\ [1mm]
         & $\pm$ 0.000 & $\pm$ 0.007 & $\pm$ 0.012 & $\pm$ 0.000 & $\pm$ 0.015 & $\pm$ 0.012 & $\pm$ 0.000 & $\pm$ 0.023 & $\pm$ 0.015 & $\pm$ 0.000 & $\pm$ 0.020 & $\pm$ 0.016 \\ [1mm]
        \hline
        \rule{0pt}{4mm} \multirow{2}{19mm}{Versicolor-Virginica} & 0.000 & 0.973 & 0.974 & 0.000 & 0.972 & 0.973  & 0.000 & 0.969  & 0.973 & 0.001 & 0.969 & 0.972  \\ [1mm]
         & $\pm$ 0.000 & $\pm$ 0.023 & $\pm$ 0.027 & $\pm$ 0.000 & $\pm$ 0.013 & $\pm$ 0.019 & $\pm$ 0.000 & $\pm$ 0.015 & $\pm$ 0.015 & $\pm$ 0.000 & $\pm$ 0.014 & $\pm$ 0.015 \\ [1mm]
         \hline
        \rule{0pt}{4mm} \multirow{2}{19mm}{Transfusion} & 0.001 & 0.973 & 0.953 & 0.003 & 0.984 & 0.966  & 0.003 & 0.990  & 0.971 & 0.003 & 0.989 & 0.974  \\ [1mm]
         & $\pm$ 0.000 & $\pm$ 0.007 & $\pm$ 0.014 & $\pm$ 0.002 & $\pm$ 0.005 & $\pm$ 0.009 & $\pm$ 0.001 & $\pm$ 0.005 & $\pm$ 0.008 & $\pm$ 0.001 & $\pm$ 0.005 & $\pm$ 0.008 \\ [1mm]
         \hline
        \rule{0pt}{4mm} \multirow{2}{19mm}{Vertebral Column} & 0.001 & 0.991 & 0.993 & 0.003 & 0.986 & 0.991  & 0.000 & 0.982  & 0.989 & 0.000 & 0.977 & 0.985  \\ [1mm]
         & $\pm$ 0.000 & $\pm$ 0.004 & $\pm$ 0.005 & $\pm$ 0.001 & $\pm$ 0.008 & $\pm$ 0.005 & $\pm$ 0.000 & $\pm$ 0.008 & $\pm$ 0.005 & $\pm$ 0.000 & $\pm$ 0.009 & $\pm$ 0.005 \\ [1mm]
         \hline
        \rule{0pt}{4mm} \multirow{2}{19mm}{Ecoli} & 0.001 & 0.997 & 0.994 & 0.000 & 0.996 & 0.995  & 0.000 & 0.997  & 0.996 & 0.000 & 0.998 & 0.996  \\ [1mm]
         & $\pm$ 0.000 & $\pm$ 0.006 & $\pm$ 0.011 & $\pm$ 0.000 & $\pm$ 0.005 & $\pm$ 0.008 & $\pm$ 0.000 & $\pm$ 0.004 & $\pm$ 0.007 & $\pm$ 0.000 & $\pm$ 0.003 & $\pm$ 0.006 \\ [1mm]
         \hline
        \rule{0pt}{4mm} \multirow{2}{19mm}{Glass} & \textbf{0.034} & \textbf{0.934} & \textbf{0.926} & \textbf{0.017} & \textbf{0.932} & \textbf{0.921}  & \textbf{0.027} & \textbf{0.926}  & \textbf{0.927} & \textbf{0.010} & \textbf{0.925} & \textbf{0.926}  \\ [1mm]
         & \textbf{$\pm$ 0.023} & \textbf{$\pm$ 0.034} & \textbf{$\pm$ 0.050} & \textbf{$\pm$ 0.002} & \textbf{$\pm$ 0.032} & \textbf{$\pm$ 0.042} & \textbf{$\pm$ 0.013} & \textbf{$\pm$ 0.031} & \textbf{$\pm$ 0.037} & \textbf{$\pm$ 0.003} & \textbf{$\pm$ 0.034} & \textbf{$\pm$ 0.034} \\ [1mm]
         \hline
        \rule{0pt}{4mm} \multirow{2}{19mm}{Breast Cancer} & 0.000 & 0.991 & 0.992 & 0.000 & 0.995 & 0.995  & 0.000 & 0.994  & 0.995 & 0.000 & 0.994 & 0.995  \\
         & $\pm$ 0.000 & $\pm$ 0.007 & $\pm$ 0.010 & $\pm$ 0.000 & $\pm$ 0.007 & $\pm$ 0.009 & $\pm$ 0.000 & $\pm$ 0.007 & $\pm$ 0.009 & $\pm$ 0.000 & $\pm$ 0.006 & $\pm$ 0.008 \\ [1mm]
         \hline
    \end{tabular}}
    \caption{The mean difference in accuracy ($\Delta\ Acc$) between the classical KNN computation and the ideal statevector QKNN simulation with its standard deviation, and the mean Jaccard Index (JI) and Average JI (AJI) with standard deviation using the KNN results computed via binary arithmetic as truth.}
    \label{tab:classical_statevector}
\end{table}

The results in Table \ref{tab:classical_statevector} compare the performance of QKNN using map-encoded data on Qiskit's \cite{Qiskit} \href{https://qiskit.github.io/qiskit-aer/stubs/qiskit_aer.StatevectorSimulator.html}{StatevectorSimulator} to the ground truth of KNN performed using only traditional binary arithmetic. It shows the mean difference in accuracy, mean $JI$, and mean $AJI$ across all 5 cross-validation folds for each dataset with their standard deviation. Besides a few exceptions from the glass dataset, the accuracy is effectively the same as the classical result and the $JI$ and $AJI$ are all nearly unity. This supports the conclusion that the map encoding method accurately computes Euclidean distance. This satisfies the first goal of this algorithm.

The edge cases can be attributed to the high sensitivity of some of the datasets, in which small differences in computation result in a different label for a test feature. The statevector result is implemented via the product of many matrices, while the classical result is obtained through relatively few arithmetic operations. The large number of operations in the statevector solution increases computational error propagation, resulting in slightly different results than the classical solution.

\begin{figure}[htb]
\centering	
\subfloat[]{
\label{subfig:statevector_aer_accuracy}
\includegraphics [width=0.33\textwidth]{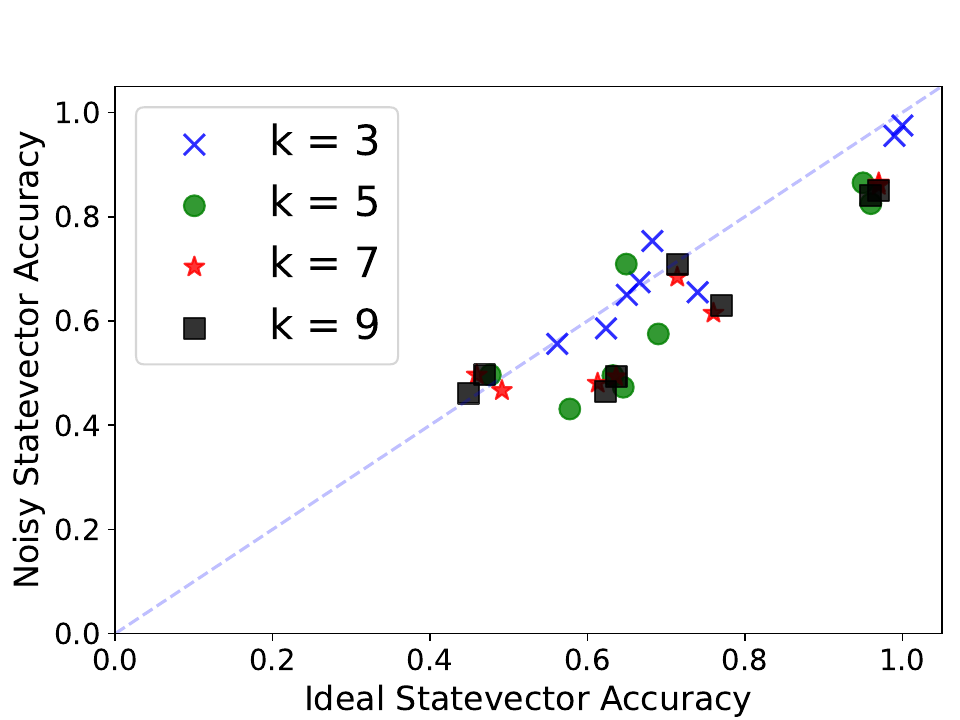}}
\subfloat[]{
\label{subfig:statevector_aer_ji}
\includegraphics [width=0.33\textwidth]{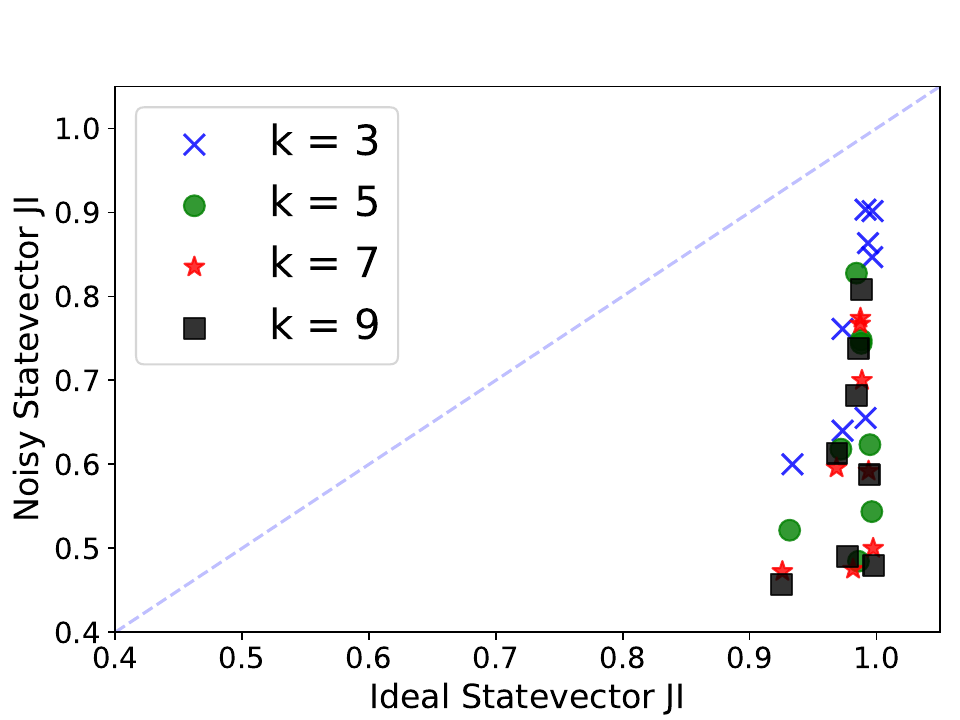}}
\subfloat[]{
\label{subfig:statevector_aer_aji}
\includegraphics [width=0.33\textwidth]{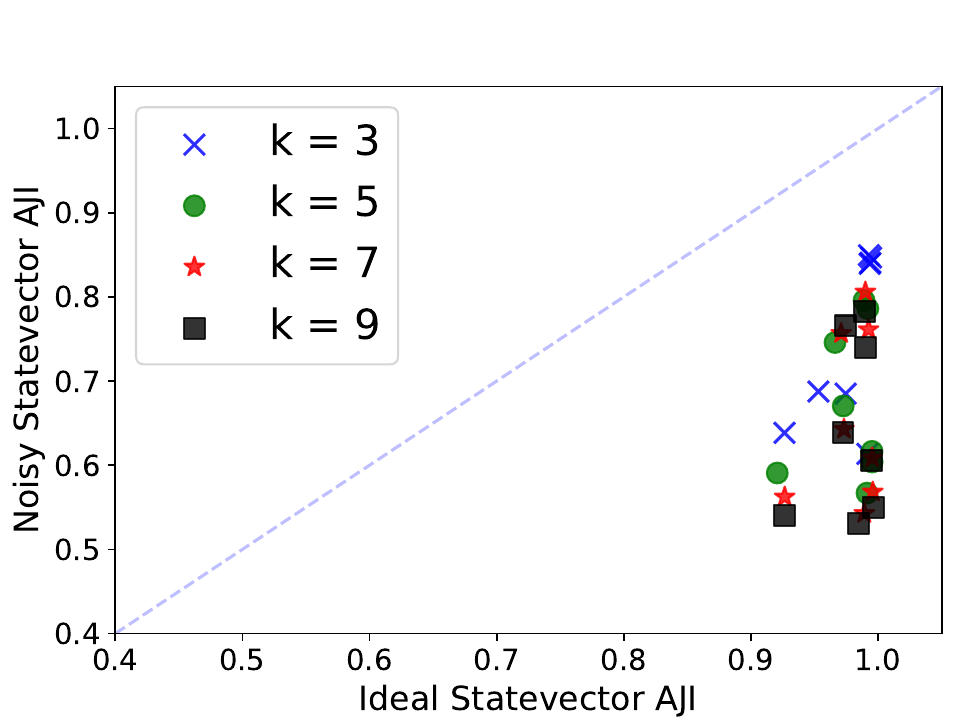}}
\caption{The mean accuracy (\ref{subfig:statevector_aer_accuracy}), $JI$ (\ref{subfig:statevector_aer_ji}), and $AJI$ (\ref{subfig:statevector_aer_aji}) of the map encoding QKNN algorithm executed on Qiskit's ideal StatevectorSimulator backend vs the same algorithm executed on Qiskit's AerSimulator backend with realistic noise.}
\label{fig:statevector_aer}
\end{figure}

The algorithm was also run on a more realistic backend - the AerSimulator from Qiskit \cite{Qiskit} using a statevector simulation with IBMQ's Guadalupe as the noise model. This backend adds noise to generate a more accurate representation of what running this experiment on a real quantum computer would be like. All runs were performed with 1024 shots. As seen in Figure \ref{fig:statevector_aer}, the performance of the Aer simulation is significantly worse than that of the nearly ideal statevector backend. With a  measurement resolution as coarse as $\frac{1}{1024}$, this poor performance is expected from this low-shots simulation. Ultimately, these shots must be distributed among the number of possible distinct states in the circuit. Using the definitions from Section \ref{sec:complexity}, this is $2^{\lceil \log_2(Q) \rceil + \lceil \log_2(N + 1) \rceil + 1}$ in our circuit.

\subsubsection{Comparison with Leading Algorithm} \label{sec:leading}

\begin{figure}[htb]
\centering	
\subfloat[]{
\label{subfig:zardini_aer_accuracy}
\includegraphics [width=0.33\textwidth]{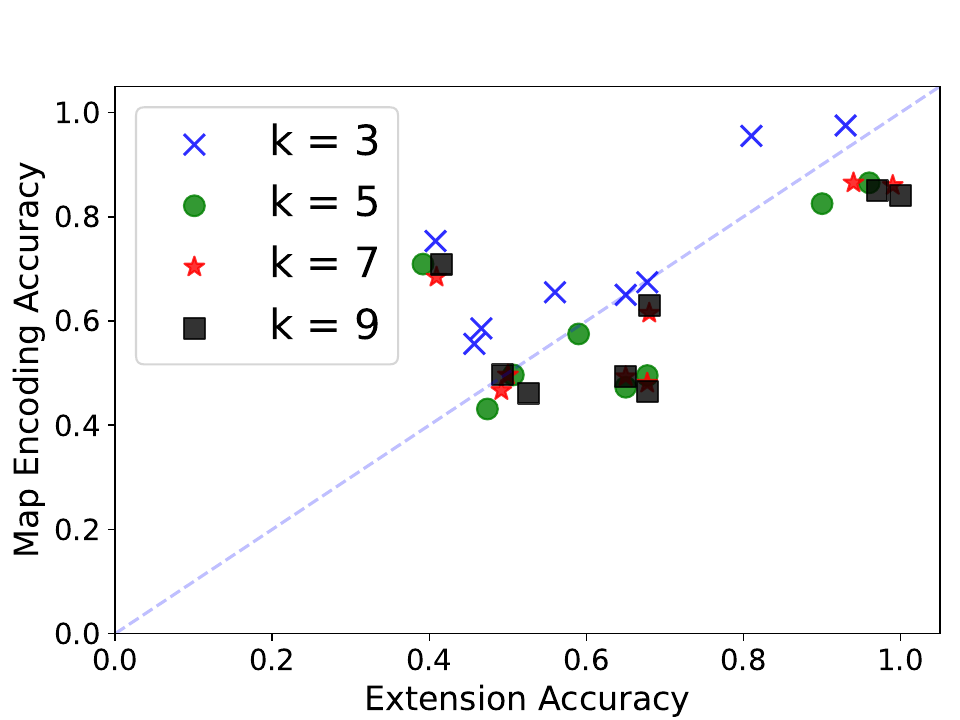}}
\subfloat[]{
\label{subfig:zardini_aer_ji}
\includegraphics [width=0.33\textwidth]{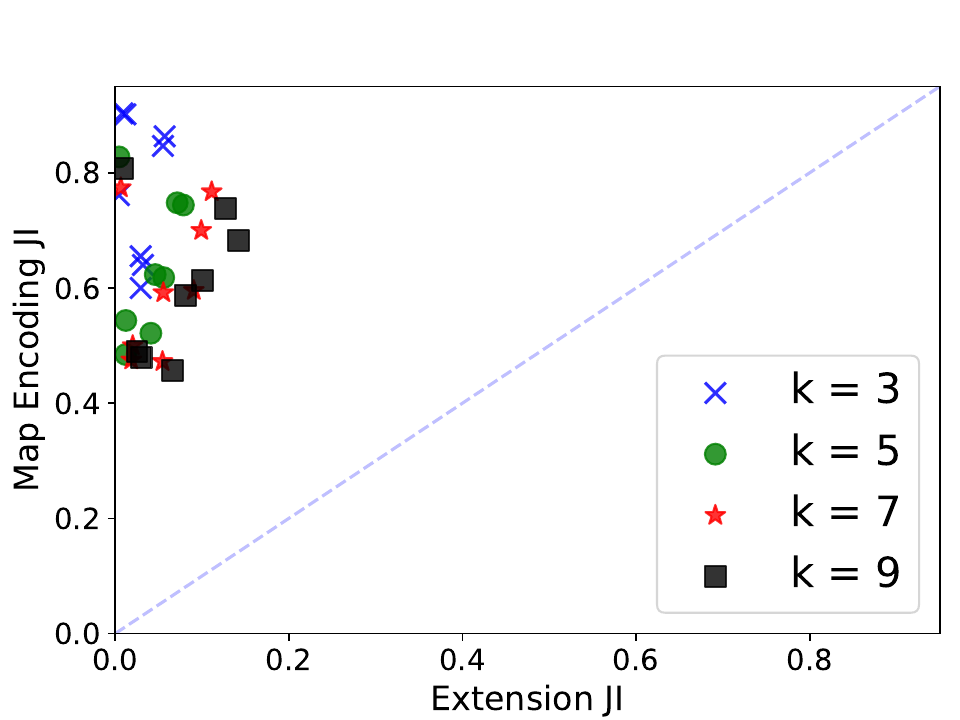}}
\subfloat[]{
\label{subfig:zardini_aer_aji}
\includegraphics [width=0.33\textwidth]{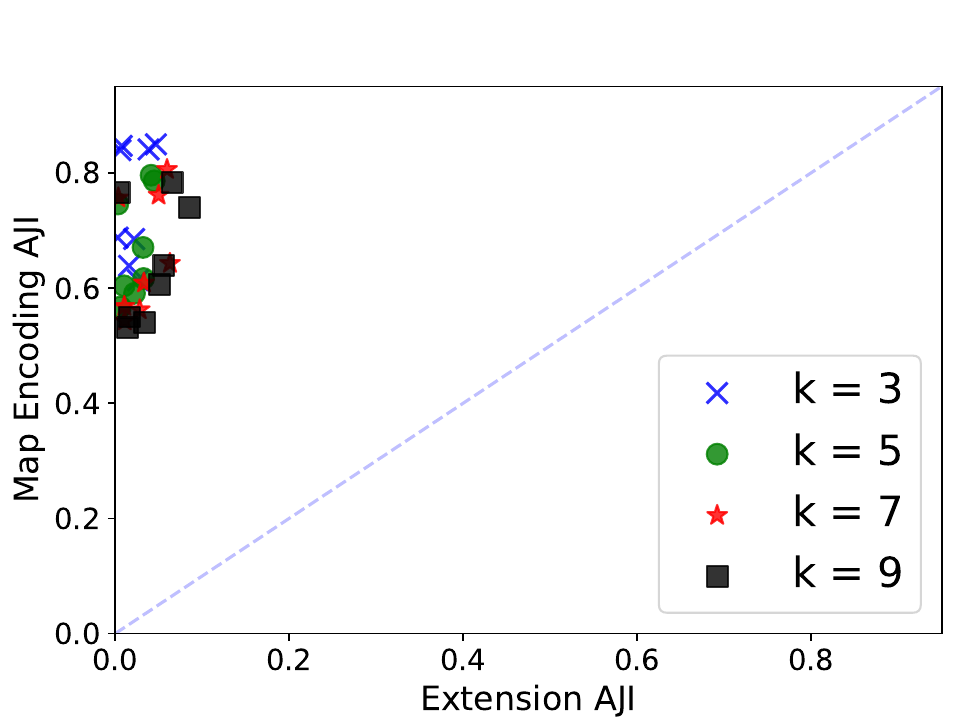}}
\caption{The mean accuracy (\ref{subfig:zardini_aer_accuracy}), $JI$ (\ref{subfig:zardini_aer_ji}), and $AJI$ (\ref{subfig:zardini_aer_aji}) of simulated QKNN using the map encoding technique compared to the extension encoding technique \cite{Zardini2023AQK}.}
\label{fig:zardini_aer}
\end{figure}

The map encoding technique is compared to the extension encoding technique \cite{Zardini2023AQK} in Figure \ref{fig:zardini_aer}. Since the accuracy metric is largely dependent on the underlying KNN algorithm, the more relevant metrics are the $JI$ and $AJI$. These metrics describe how closely the quantum results mirror the classical results. As such, it is clear that the map encoding technique replicates the classical KNN much better than the extension encoding technique in a realistic simulation. 

There could be a few possible explanations for the difference in performance. While both methods used Qiskit's AerSimulator, minor differences in implementation could explain some variations. However, the result of these differences is expected to be minimal. A more plausible explanation for the large difference in performance is the circuit width. Our method uses $\lceil \log_2(\frac{2N + 3}{N + 1}) \rceil + 1$ fewer qubits, which allows the KNN step to make much finer label selections for test samples.

\subsubsection{Performance with More Shots}

\begin{figure}[htb]
\centering
\includegraphics [width=0.8\textwidth]{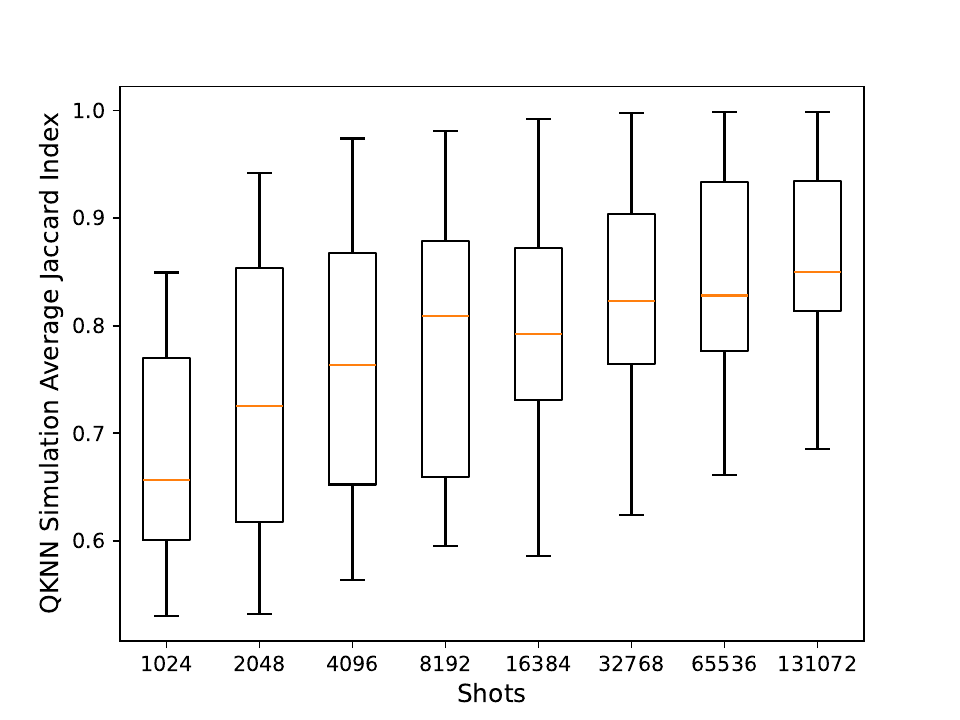}
\caption{The effect of increasing the number of circuit executions and measurements (shots) on the Average Jaccard Index}
\label{fig:shots}
\end{figure}

It is expected that the integrity of the quantum result with respect to the classical result increases as the number of shots increases because the resolution of the measurement is inversely proportional to the number of shots. To study this, the map-encoded QKNN was executed on Qiskit's \cite{Qiskit} AerSimulator with increasing shots. The results can be viewed in Figure \ref{fig:shots}. As expected, the $AJI$ increases as shots increase, providing evidence for the claim of positive correlation between shots and integrity. It is worth noting that the standard deviation first increases and then decreases. This indicates that the algorithm consistently underperforms at lower shot counts due to low fidelity, reaches a point at which the results become meaningful to the KNN algorithm, and then becomes more reliable with increasing shots.

\section{Conclusions and Future Work} \label{sec:conclusions}

This report described a practical QKNN implementation that used a new map encoding technique to load data into quantum computing hardware. This novel technique allows for efficient comparison of sets of vectors with different $L2$ norm values. This algorithm required fewer qubits than the extension and translation encoding \cite{Zardini2023AQK}. It is less complex to implement and arguably easier to understand than these other methods. It also computes the true Euclidean distance, which can be useful in contexts with geometrical implications. The realization of QRAM in the future would considerably decrease the circuit depth and execution speed. It would also increase fidelity due to less decoherence \cite{decoherence}. 

An implementation of this algorithm has been provided in a public \href{https://github.com/crect500/squib}{repo}, along with the performance results. It has been shown in Section \ref{sec:performance} that the implementation accurately models the classical implementation. In Section \ref{sec:leading}, this algorithm demonstrated superior performance to the algorithm presented in \cite{Zardini2023AQK} under similar operating conditions.

Regarding future applications of this method, several use cases are immediately attractive. For example, this algorithm would be much better suited to the k-means algorithm \cite{Yadav2013ARO}, as it can store the distances between all centroids and all features in the same superposition of states. Using KNN as a case study for map encoding was due to the larger volume and higher quality of the publications available on QKNN. This allowed for a more relevant evaluation of the new algorithm, the success of which can now be applied to other applications.

\section{Funding}

This research was performed without the receipt of grants or any other funding.

\bibliography{references}

@article{Shor1995PolynomialTimeAF,
  title={Polynomial-Time Algorithms for Prime Factorization and Discrete Logarithms on a Quantum Computer},
  author={Peter W. Shor},
  journal={SIAM Rev.},
  year={1999},
  volume={41},
  pages={303-332},
  url={https://api.semanticscholar.org/CorpusID:2337707}
}

@article{Preskill2012QuantumCA,
  title={Quantum computing and the entanglement frontier},
  author={John Preskill},
  journal={arXiv:1203.5813},
  year={2012},
  url={https://api.semanticscholar.org/CorpusID:48828406}
}

@article{Grover1997QuantumMH,
  title={Quantum Mechanics Helps in Searching for a Needle in a Haystack},
  author={Lov K. Grover},
  journal={Physical Review Letters},
  year={1997},
  volume={79},
  pages={325-328},
  url={https://api.semanticscholar.org/CorpusID:1621022}
}

@article{Hur2021QuantumCN,
  title={Quantum convolutional neural network for classical data classification},
  author={Tak Hur and Lee Yeong Kim and Daniel Kyungdeock Park},
  journal={Quantum Machine Intelligence},
  year={2021},
  volume={4},
  url={https://api.semanticscholar.org/CorpusID:236772493}
}

@article{Venturelli2018ReverseQA,
  title={Reverse quantum annealing approach to portfolio optimization problems},
  author={Davide Venturelli and Alex Kondratyev},
  journal={Quantum Machine Intelligence},
  year={2018},
  volume={1},
  pages={17 - 30},
  url={https://api.semanticscholar.org/CorpusID:53559618}
}

@article{Simon2023OnNQ,
  title={On Neural Quantum Support Vector Machines},
  author={Lars Simon and Manuel Radons},
  journal={Quantum Mach. Intell.},
  year={2023},
  volume={6},
  pages={3},
  url={https://api.semanticscholar.org/CorpusID:260926604}
}

@article{Mohanty2023SolvingTV,
  title={Solving the vehicle routing problem via quantum support vector machines},
  author={Nishikanta Mohanty and Bikash K. Behera and Christopher Ferrie},
  journal={Quantum Mach. Intell.},
  year={2023},
  volume={6},
  pages={34},
  url={https://api.semanticscholar.org/CorpusID:260735494}
}

@article{Wu2021QuantumKA,
  title={Quantum k-means algorithm based on Manhattan distance},
  author={Zhihao Wu and Tingting Song and Yanbing Zhang},
  journal={Quantum Information Processing},
  year={2021},
  volume={21},
  url={https://api.semanticscholar.org/CorpusID:245673498}
}

@article{Ohno2022AQA,
  title={A quantum algorithm of K-means toward practical use},
  author={Hiroshi Ohno},
  journal={Quantum Information Processing},
  year={2022},
  volume={21},
  url={https://api.semanticscholar.org/CorpusID:247989001}
}

@article{Wiebe2014QuantumNA,
  title={Quantum Nearest-Neighbor Algorithms for Machine Learning},
  author={Nathan Wiebe and Ashish Kapoor and Krysta Marie Svore},
  journal={Quantum Information \& Computation},
  year={2014},
  volume={15},
  pages={318-358},
  url={https://api.semanticscholar.org/CorpusID:37339559}
}

@article{fix1985discriminatory,
  title={Discriminatory analysis: nonparametric discrimination, consistency properties},
  author={Fix, Evelyn},
  volume={57},
  pages={238-247},
  journal={International Statistical Review},
  year={1985},
  publisher={USAF school of Aviation Medicine}
}

@article{MaldonadoRomo2024QuantumKN,
  title={Quantum K-Nearest Neighbors: Utilizing QRAM and SWAP-Test Techniques for Enhanced Performance},
  author={Alberto Maldonado-Romo and Jes{\'u}s Yalj{\'a} Montiel-P{\'e}rez and Victor Onofre and Javier Maldonado-Romo and Juan Humberto Sossa-Azuela},
  journal={Mathematics},
  year={2024},
  url={https://api.semanticscholar.org/CorpusID:270578631}
}

@article{Ronggon2026QKNNNQ,
  title={QKNN: Noise‐Resilient Quantum KNN Algorithm for High‐Accuracy Classification},
  author={Asif Akhtab Ronggon and Tuhin Hossain and Tahani Jaser Alahmadi and Mohammed Ali Moni},
  journal={Advanced Quantum Technologies},
  year={2026},
  volume={9},
  url={https://api.semanticscholar.org/CorpusID:285229277}
}

@article{Zardini2023AQK,
  title={A quantum k-nearest neighbors algorithm based on the Euclidean distance estimation},
  author={Enrico Zardini and Enrico Blanzieri and Davide Pastorello},
  journal={Quantum Mach. Intell.},
  year={2023},
  volume={6},
  pages={23},
  url={https://api.semanticscholar.org/CorpusID:258557902}
}

@article{Tessari2024SurpassingCS,
  title={Surpassing Cosine Similarity for Multidimensional Comparisons: Dimension Insensitive Euclidean Metric (DIEM)},
  author={Federico Tessari and Neville Hogan},
  journal={ArXiv},
  year={2024},
  volume={abs/2407.08623},
  url={https://api.semanticscholar.org/CorpusID:271097701}
}

@inproceedings{Gao2022QuantumKN,
  title={Quantum K-nearest neighbors classification algorithm based on Mahalanobis distance},
  author={Li-Zhen Gao and Chun-Yue Lu and Gongde Guo and Xin Zhang and Song Lin},
  booktitle={Frontiers of Physics},
  year={2022},
  url={https://api.semanticscholar.org/CorpusID:253069511}
}

@article{Ma2021QuantumAF,
  title={Quantum Algorithm for K-Nearest Neighbors Classification Based on the Categorical Tensor Network States},
  author={Yanping Ma and Hongqing Song and Jun Zhang},
  journal={International Journal of Theoretical Physics},
  year={2021},
  volume={60},
  pages={1164 - 1174},
  url={https://api.semanticscholar.org/CorpusID:233783361}
}

@article{Yuan2017QuantumIF,
  title={Quantum Image Filtering in the Spatial Domain},
  author={Suzhen Yuan and Xuefeng Mao and Jing Zhou and Xiaofa Wang},
  journal={International Journal of Theoretical Physics},
  year={2017},
  volume={56},
  pages={2495 - 2511},
  url={https://api.semanticscholar.org/CorpusID:126372336}
}

@inproceedings{Nielsen2011QuantumCA,
  title={Quantum Computation and Quantum Information (10th Anniversary edition)},
  author={Michael A. Nielsen and Isaac L. Chuang},
  year={2011},
  url={https://api.semanticscholar.org/CorpusID:263408726}
}

@article{Ruan2017QuantumAF,
  title={Quantum Algorithm for K-Nearest Neighbors Classification Based on the Metric of Hamming Distance},
  author={Yue Ruan and Xiling Xue and Heng Liu and Jianing Tan and Xi Li},
  journal={International Journal of Theoretical Physics},
  year={2017},
  volume={56},
  pages={3496 - 3507},
  url={https://api.semanticscholar.org/CorpusID:125493100}
}

@article{Zhou2021QuantumKI,
  title={Quantum K-Nearest-Neighbor Image Classification Algorithm Based on K-L Transform},
  author={Nanrun Zhou and Xiumin Liu and Yu-Ling Chen and Nisuo Du},
  journal={International Journal of Theoretical Physics},
  year={2021},
  volume={60},
  pages={1209 - 1224},
  url={https://api.semanticscholar.org/CorpusID:233681660}
}

@article{Zhang2013NEQRAN,
  title={NEQR: a novel enhanced quantum representation of digital images},
  author={Yi Zhang and Kai Lu and Yinghui Gao and Mo Wang},
  journal={Quantum Information Processing},
  year={2013},
  volume={12},
  pages={2833 - 2860},
  url={https://api.semanticscholar.org/CorpusID:254977084}
}

@article{Li2021QuantumKN,
  title={Quantum K-nearest neighbor classification algorithm based on Hamming distance},
  author={Jing Li and Song Lin and Kai-huan Yu and Gongde Guo},
  journal={Quantum Information Processing},
  year={2021},
  volume={21},
  url={https://api.semanticscholar.org/CorpusID:232147885}
}

@article{Phalak2023QuantumRA,
  title={Quantum Random Access Memory for Dummies},
  author={Koustubh Phalak and Avimita Chatterjee and Swaroop Ghosh},
  journal={Sensors (Basel, Switzerland)},
  year={2023},
  volume={23},
  url={https://api.semanticscholar.org/CorpusID:258436702}
}

@article{Brassard2000QuantumAA,
  title={Quantum Amplitude Amplification and Estimation},
  author={Gilles Brassard and Peter H{\o}yer and Michele Mosca and A. Montreal and Brics University of Aarhus and Cacr University of Waterloo},
  journal={arXiv: Quantum Physics},
  year={2000},
  url={https://api.semanticscholar.org/CorpusID:54753}
}

@article{Hai2022NewAO,
  title={New approach of KNN Algorithm in quantum computing based on new design of quantum circuits},
  author={Vu Tuan Hai and Phan Hoang Chuong and Pham The Bao},
  journal={Informatica (Slovenia)},
  year={2022},
  volume={46},
  url={https://api.semanticscholar.org/CorpusID:248289195}
}

@inproceedings{Yadav2013ARO,
  title={A Review of K-mean Algorithm},
  author={Jyoti Yadav and Monika Sharma},
  year={2013},
  url={https://api.semanticscholar.org/CorpusID:61532973}
}

@article{Afham2020QuantumKN,
  title={Quantum k-nearest neighbor machine learning algorithm},
  author={Afham and Afrad Basheer and Sandeep K. Goyal},
  journal={arXiv: Quantum Physics},
  year={2020},
  url={https://api.semanticscholar.org/CorpusID:214605675}
}

@article{LunaMrquez2025kNNCU,
  title={kNN classification using Hausdorff metric with quantum-based Euclidean distance},
  author={Diego Carlos Luna-M{\'a}rquez and Ra{\'u}l Pinto-El{\'i}as and Tom{\'a}s P{\'e}rez-Becerra and Andrea Magad{\'a}n Salazar and Nimrod Gonz{\'a}lez-Franco and Jorge Fuentes-Pacheco},
  journal={Quantum Mach. Intell.},
  year={2025},
  volume={7},
  pages={54},
  url={https://api.semanticscholar.org/CorpusID:277875228}
}

@article{Yu2020,
author = {Yu, Kai and Guo, Gong-De and Li, Jing and Lin, Song},
year = {2020},
month = {10},
pages = {},
title = {Quantum Algorithms for Similarity Measurement Based on Euclidean Distance},
volume = {59},
journal = {International Journal of Theoretical Physics},
doi = {10.1007/s10773-020-04567-1}
}

@article{Zindorf2024EfficientIO,
  title={Efficient implementation of multicontrolled quantum gates},
  author={Ben Zindorf and Sougato Bose},
  journal={Physical Review Applied},
  year={2024},
  url={https://api.semanticscholar.org/CorpusID:268875935}
}

@article{Giovannetti2007QuantumRA,
  title={Quantum random access memory.},
  author={Vittorio Giovannetti and Seth Lloyd and Lorenzo Maccone},
  journal={Physical review letters},
  year={2007},
  volume={100 16},
  pages={
          160501
        },
  url={https://api.semanticscholar.org/CorpusID:570390}
}

@article{Araujo2020ADA,
  title={A divide-and-conquer algorithm for quantum state preparation},
  author={Israel F. Araujo and Daniel Kyungdeock Park and Francesco Petruccione and Adenilton Jos{\'e} da Silva},
  journal={Scientific Reports},
  year={2020},
  volume={11},
  url={https://api.semanticscholar.org/CorpusID:220961607}
}

@misc{Qiskit,
    author = {{Qiskit contributors}},
    title = {Qiskit: An Open-source Framework for Quantum Computing},
    year = {2023},
    doi = {10.5281/zenodo.2573505}
}

@inproceedings{Fastovets2019MachineLM,
  title={Machine learning methods in quantum computing theory},
  author={Dmitrii V. Fastovets and Yu. I. Bogdanov and Boris I. Bantysh and V F Lukichev},
  booktitle={International Conference on Micro- and Nano-Electronics},
  year={2019},
  url={https://api.semanticscholar.org/CorpusID:88499935}
}

@article{
decoherence,
author = {I. L. Chuang  and R. Laflamme  and P. W. Shor  and W. H. Zurek },
title = {Quantum Computers, Factoring, and Decoherence},
journal = {Science},
volume = {270},
number = {5242},
pages = {1633-1635},
year = {1995},
doi = {10.1126/science.270.5242.1633},
URL = {https://www.science.org/doi/abs/10.1126/science.270.5242.1633},
eprint = {https://www.science.org/doi/pdf/10.1126/science.270.5242.1633}
}

@article{Schuld2017DistanceBased,
  title={Implementing a distance-based classifier with a quantum interference circuit},
  author={Maria Schuld and Mark Fingerhuth and Francesco Petruccione},
  journal={Europhysics Letters},
  volume={119},
  pages={60002},
  year={2017},
  doi={10.1209/0295-5075/119/60002}
}

@article{Berti2024RoleOE,
  title={The role of encodings and distance metrics for the quantum nearest neighbor},
  author={Alessandro Berti and Anna Bernasconi and Del Corso, Gianna M. and Riccardo Guidotti},
  journal={Quantum Mach. Intell.},
  volume={6},
  pages={62},
  year={2024},
  doi={10.1007/s42484-024-00197-6}
}

@article{Lloyd2013QuantumAF,
  title={Quantum algorithms for supervised and unsupervised machine learning},
  author={Seth Lloyd and Masoud Mohseni and Patrick Rebentrost},
  journal={arXiv:1307.0411},
  year={2013}
}

@inproceedings{Bachrach2014SpeedingUT,
  title={Speeding up the {Xbox} recommender system using a {Euclidean} transformation for inner-product spaces},
  author={Yoram Bachrach and Yehuda Finkelstein and Ran Gilad-Bachrach and Liran Katzir and Noam Koenigstein and Nir Nice and Ulrich Paquet},
  booktitle={Proceedings of the 8th ACM Conference on Recommender Systems},
  pages={257--264},
  year={2014},
  doi={10.1145/2645710.2645741}
}

\end{document}